\documentclass{article}
\usepackage{spconf,amsmath,graphicx,booktabs,multirow,amsfonts,xcolor,hyperref}

\title{PIT-QMM: A Large Multimodal Model For No-Reference Point Cloud Quality Assessment}
\name{Shashank Gupta$^{\star \dagger}$ \qquad Gregoire Phillips$^{\dagger}$ \qquad Alan C Bovik$^{\star}$\thanks{Work done during an internship at Ericsson Research Silicon Valley.}}
\address{$^{\star}$ The University of Texas at Austin \\
      $^{\dagger}$Ericsson Research}
\begin{document}
%
\maketitle
\begin{abstract}
Large Multimodal Models (LMMs) have recently enabled considerable advances in the realm of image and video quality assessment, but this progress has yet to be fully explored in the domain of 3D assets. We are interested in using these models to conduct No-Reference Point Cloud Quality Assessment (NR-PCQA), where the aim is to automatically evaluate the perceptual quality of a point cloud in absence of a reference. We begin with the observation that different modalities of data – text descriptions, 2D projections, and 3D point cloud views – provide complementary information about point cloud quality. We then construct PIT-QMM, a novel LMM for NR-PCQA that is capable of consuming text, images and point clouds end-to-end to predict quality scores.  Extensive experimentation shows that our proposed method outperforms the state-of-the-art by significant margins on popular benchmarks with fewer training iterations. We also demonstrate that our framework enables distortion localization and identification, which paves a new way forward for model explainability and interactivity. Code and datasets are available at \href{https://www.github.com/shngt/pit-qmm}{https://www.github.com/shngt/pit-qmm}.
\end{abstract}
\section{Introduction}
\label{sec:intro}

Point clouds, collections of 3D points with attributes like color and opacity, are fundamental to applications such as autonomous driving, immersive gaming, and digital twins \cite{zhang2022pointclip}. Their flexibility allows detailed spatial analysis with minimal geometric assumptions but makes them susceptible to distortions from sensor inaccuracies, compression, and transmission errors, which degrade perceptual quality and impair downstream tasks.


To address this, automated point cloud quality assessment (PCQA) has become a critical research focus. 
Traditional metrics like PSNR and SSIM~\cite{wang2004image}, adapted from image/video quality assessment, fail to capture the complexities of 3D data. 
Learning-based methods are also not that effective, as most PCQA datasets contain only a few hundred samples. 


Recently, large multimodal models (LMMs) trained on vast datasets have set benchmarks in 2D quality assessment. However, they are not easily extendable to the 3D case. 
Point-text multimodal models have been developed for semantic tasks such as object classification.
However, due to computational constraints, they are restricted to smaller point clouds, for which the quality problem is no longer meaningful. 
Thus, while image-text models excel in quality assessment and point-text models in 3D comprehension, neither fully captures both aspects needed for PCQA.


To bridge this gap, we propose the Point-Image-Text Quality Multimodal Model (PIT-QMM), the first end-to-end point-image-text LMM for PCQA. 
PIT-QMM leverages complementary strengths of multiple modalities:
PIT-QMM leverages the complementary strengths of different modalities: point cloud patches capture local variations often lost in 2D projections, image projections provide a global perspective, and text inputs add psychometric context and priming for the quality task. 
LMMs also excel in visual localization -- linking specific regions with textual cues -- which PIT-QMM leverages to accurately localize and categorize quality issues.

Our main contributions may be summarized as follows:

\begin{itemize}
    \item We propose PIT-QMM, the first end-to-end point-image-text multimodal model tailored for PCQA. We also introduce task-aware prompts, efficient encoder-aware point cloud sampling, and a two-stage training strategy for effective multimodal fusion as an enhancement over prior work.
    \item We perform thorough benchmarking, and show that our model beats state-of-the-art (SOTA) methods by a large margin with fewer training iterations. We validate the importance of each modality with thorough ablations.
    \item We show that PIT-QMM can identify specific distortions and their locations when prompted. Not only does this enhance interpretability and overall utility, it hints at potential reasoning capabilities about quality. 
    To our knowledge, this is the first exploration of quality localization in the point cloud domain.
\end{itemize}

Supplementary material is available at\\
\href{https://dx.doi.org/10.60864/6kge-6c07}{https://dx.doi.org/10.60864/6kge-6c07}.


\begin{figure*}
    \centering
    \includegraphics[width=0.75\linewidth]{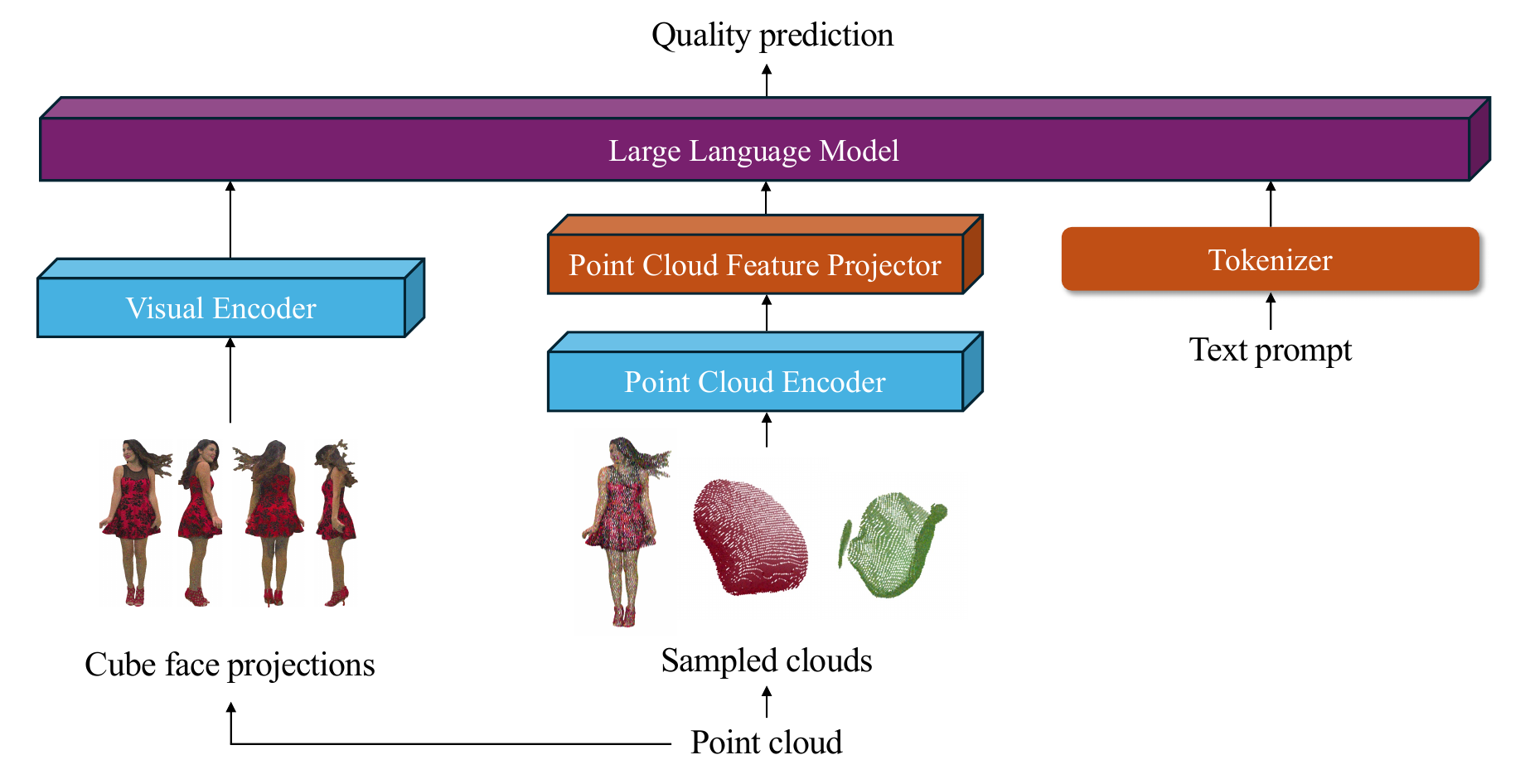}
    \caption{\textbf{An overview of the proposed Point-Image-Text Quality Multimodal Model (PIT-QMM).} PIT-QMM takes a raw point cloud and extracts both 2D and 3D views. Rich feature representations of these views are encoded by pretrained foundation models. These representations are then passed into a large multimodal model along with a textual description of the task and experimental setup, which is trained to predict quality scores.}
    \label{fig:PIT-QMM-arch}
\end{figure*}
\section{Related Work}
\label{related_work}

Traditional NR-PCQA models involve handcrafted features and regressors, which have limited expressiveness. Deep learning-based methods, such as ResSCNN~\cite{liu2023point} and MM-PCQA~\cite{zhang2022mm}, leverage 3D neural architectures for PCQA but struggle to generalize.
CoPA~\cite{shan2024contrastive} leverages the large-scale LS-PCQA \cite{liu2023point} to employ contrastive learning to obtain robust quality features, but has high pre-training costs. 


LMMs like Q-Align~\cite{wu2023q} achieve state-of-the-art results in 2D domains, but extending them to 3D by using 2D projections of the content does not work well, due to loss of local variations, occluded sections, and depth ambiguity. 
While methods like LMM-PCQA~\cite{zhang2024lmm} integrate 2D LMM predictions with handcrafted cloud features, they do not leverage expressive learned representations or multimodal interactions. Moreover, extracting the handcrafted features involves expensive k-NN-based preprocessing on the entire cloud, making it impractical for large clouds.

PIT-QMM overcomes these issues by integrating deep point cloud encoders with existing image-text LMMs, enabling end-to-end multimodal training that fully exploits the complementary strengths of point cloud, image, and text modalities for superior PCQA performance.



\section{Method}
\label{method}

This section outlines the construction of our instruction-following dataset, the architecture of PIT-QMM, and our experimental setup and model modifications for distortion localization and identification. Section~\ref{instruct_data} addresses the dataset construction, while Section~\ref{model_architecture} details how the modalities are encoded and processed to produce the desired output.


\subsection{Point-Image-Text Instruction-Following Quality Data}
\label{instruct_data}

\subsubsection{Point Clouds}
\label{pc_input}

A key challenge in including point clouds is their large size in quality assessment datasets. 
Popular encoders like Point-BERT~\cite{Point-BERT} are pre-trained on point clouds containing thousands of points, whereas quality assessment datasets feature millions, making direct input infeasible. Therefore, the point cloud is subsampled to capture multiple levels of information.

First, furthest point sampling is applied to create a sparse global view, capturing overall shape and content-level attributes.
Next, small local patches are randomly sampled to detect high-frequency local distortions. To construct each patch, we randomly select an anchor point and take its k-nearest neighbours.  
We also explored a two-scale variant which combines patches from the original and a downsampled cloud for multi-level granularity.

It is important to note that altogether these samples comprise only 3-5\% of the total cloud. 
While this allows high sample efficiency and inference speed, they are not a holistic representation, necessitating a complementary global view.


\subsubsection{Image Projections}

To address the limitations of point cloud sampling, we incorporate multi-view image projections.
For a point cloud $P$, we normalize it to zero-mean and unit-maximum distance with $\mathcal{N}(\cdot)$, then render $\mathcal{N}(P)$ into multi-view images $\{x_i \in \mathbb{R}^{H\times W\times C} |^{6}_{i=1}\}$ from six perpendicular viewpoints (i.e., along the positive and negative directions of the $x, y, z$-axes) with fixed viewing distances.
Where point cloud features provide local quality perspectives, these image projections provide a global quality perspective and allow leveraging pretrained image quality models.
\vspace{-10pt}
\subsubsection{Text}


The textual component of the dataset primes the model for no-reference quality assessment against single-stimulus absolute category ratings, conveying psychometric context and driving it to draw on relevant world knowledge.
Point cloud rendering parameters like point size and viewing distance, which influence quality, are also encoded in the prompt.

\vspace{-10pt}
\subsubsection{Final Instruction-Following Prompt}

The final input format, as in Table~\ref{tab:input_prompt}, combines point cloud data, image projections, and text into a multi-modal question-answer structure. 
Special tokens \verb|<p_start>| and \verb|<p_end>| mark the start and end of the point cloud input. 
The model predicts discrete quality levels, as detailed in Section~\ref{labels_proc}.

\begin{table}\small
\centering
\caption{\textbf{Instruction following prompt.} 
\textcolor{orange}{\{Experimental Setup\}} describes the psychometric setup. \textcolor{violet}{\{im\_tokens\}} are image tokens and \textcolor{blue}{\{p\_tokens\}} are point tokens.}


\label{tab:input_prompt}
\scalebox{0.89}{
\begin{tabular}{@{}l@{\hspace{1em}}p{0.80\linewidth}@{}} 
\toprule
\textcolor{red}{\texttt{\{}System Prompt\texttt{\}}} & \\ \midrule
USER: & 
\begin{minipage}[t]{\linewidth}
This is a point cloud rated for quality. It was displayed to a human in a single stimulus setup with absolute category ratings. \textcolor{orange}{\{Experimental Setup\}}\textcolor{violet}{\texttt{\{}im\_tokens\texttt{\}}}\texttt{<}p\_start\texttt{>}\textcolor{blue}{\texttt{\{}p\_tokens\texttt{\}}}\texttt{<}p\_end\texttt{>} Can you rate the quality of the point cloud?
\end{minipage} \\ 
ASSISTANT: & The quality of the point cloud is excellent. \\ 
\bottomrule
\end{tabular}
}
\end{table}

\vspace{-10pt}
\subsection{Model Architecture}
\label{model_architecture}

As shown in Figure~\ref{fig:PIT-QMM-arch}, our PIT-QMM is a generative model that aims to complete multi-modal sentences containing point clouds, images and text. 
The model consists of four main components - an image encoder $f_{im}$, a point cloud encoder $f_{point}$, a point cloud embedding projector $f_{point\_proj}$, and a large language model (LLM) backbone $f_{llm}$. 

The point cloud encoder $f_{point}$ takes in a point cloud $P \in \mathbb{R}^{s\times n\times d}$, where $s$ is the number of patches, $n$ is the patch size and $d$ is the feature dimension. 
The output is a sequence of patched point features $X \in \mathbb{R}^{s\times m\times c}$, where $m$ is the number of patch features and $c$ is the feature dimension. 
The projector $f_{proj}$ is a multi-layer perceptron (MLP) that maps the point features $X$ to point tokens $Y \in \mathbb{R}^{s\times m\times c'}$, where $c'$ matches the dimension of the text and image tokens. 
Finally this is flattened to $Z \in \mathbb{R}^{sm\times c'}$, which we feed into $f_{llm}$. 

The LLM $f_{llm}$ takes in a sequence in $\mathbb{R}^{n'\times c'}$, where $n'$ is the length of the total input token sequence. As a decoder-only LLM, it produces a probability distribution for the next token of size $\mathbb{R}^V$, where $V$ is the vocabulary size.

\vspace{-10pt}
\subsection{Training and Inference}


\subsubsection{Label Smoothing and Discretization}
\label{labels_proc}

As observed in Q-Align, LMMs optimized for quality prediction perform better when they are asked to produce discrete text labels, largely due to their bias to produce text as opposed to numeric values. 
We follow a similar discretization strategy during training and convert continuous quality scores to five-point Likert levels. 
During inference, discrete outputs are mapped to continuous scores by taking a weighted average of numeric label levels based on output token probabilities.
\vspace{-10pt}

\subsubsection{Two-stage Training}



We employ a two stage training strategy. 
In the first feature alignment stage,  the parameters of the point cloud projector are trained while others remain frozen. 
This stage uses small point clouds from the Cap3D~\cite{Cap3D} dataset.
The point cloud sampling strategy is not applied here, as the input size is small.
In the second instruction-tuning stage, we unfreeze the image abstractor and add LoRA~\cite{hu2021lora} adapters to the LLM and the point cloud encoder. 
The model is fine-tuned end-to-end using the constructed quality dataset. 
During this stage, the image abstractor adapts to the domain of 2D projections, and the point cloud encoder adjusts to the domain of local patches with high-frequency variations.

\subsection{Distortion Identification and Localization}

In order to demonstrate the quality representation abilities of our model, we constructed a synthetic distortion identification and localization task. Specifically, we took pristine clouds, isolated a specific octant of each and applied a distortion from a predefined bank on it, and merged it back to the original cloud. The model is now fine-tuned to predict the octant and type of distortion from the distorted cloud.

Performing well on this task requires two key modifications.
First, since random patches may not cover all octants, we deterministically sample patches to cover each octant.
Since this may exceed the context length, we average pool the point cloud features within each patch before passing on to the LLM.
Next, since the visual tokens inherently contain no information about which projection they belong to, we add learnable position embeddings shared across tokens originating from the same view. 
This allows the model to discriminate features from different views, which aids in localization.

\section{Experiments}

\subsection{Datasets}
\label{datasets}

Our experiments are based on three popular PCQA datasets, namely LS-PCQA~\cite{liu2023point}, SJTU-PCQA~\cite{9238424}, and WPC~\cite{liu2022perceptual}. LS-PCQA is a large-scale PCQA dataset with 104 pristine and 24,024 distorted point clouds. Each pristine point cloud is impaired by 33 types of distortions at 7 levels of severity. The labels in LS-PCQA are mostly synthetically geenerated pseudo-MOSs, with only 930 samples having psychometrically-collected true MOSs. We term this subset LSPCQA-small and report results of ablations on it, along with WPC. SJTU-PCQA contains 9 reference and 378 distorted samples impaired by 7 types of distortions at 6 levels, while WPC contains 20 reference point clouds and 740 distorted samples disturbed by 5 types of distortions.

\subsection{Evaluation Protocol}

We tested PIT-QMM against other SOTA models on all datasets in Section~\ref{datasets}. 
We first constructed instruction-tuning data from the raw datasets, as in Section~\ref{instruct_data}. Each sample is thus a set of point cloud samples, cubic image projections and instruction text. 
We split each dataset into content-separated train-test sets in a 4:1 ratio. 
We minimized loss on the training set and obtained metrics on the test set.
Due to the randomness involved in sampling from the point cloud, we computed metrics on the test set with 10 different seeds and took the mean.
Finally, the test metrics were averaged over 5 different train-test splits to obtain the final reported metrics. 
Two popular evaluation metrics were used to quantify the agreement between predicted quality scores and MOSs: Spearman rank order correlation coefficient (SROCC) and Pearson linear correlation coefficient (PLCC).

\subsection{Implementation Details}



Our experiments were performed with PyTorch using $3\times 40$ GB NVIDIA A100 GPUs. 
For the point cloud encoder, we used Point-BERT pretrained with ULIP-2~\cite{xue2024ulip}. 
We sampled three patches in total, including the furthest point sample of the cloud.
The point cloud projector is a randomly initialized MLP. 
The image encoder is a Vit-L/14 and the LLM is taken from mPLUG-Owl2. 

For the alignment stage, we pretrained on the instruction-following variant of Cap3D from Point-LLM~\cite{xu2023pointllm} for 3 epochs with a batch size of 12.  
We used a learning rate of $2\times 10^{-3}$ with cosine annealing and a warmup of 0.3.
During finetuning, we trained on LS-PCQA for 5 epochs, SJTU-PCQA for 90 epochs and WPC for 30 epochs. 
We used a learning rate of $2\times 10^{-4}$ with cosine annealing and a warmup of 0.3. 
For LoRA, we used $r=128$, $\alpha=256$, and $p=0.05$ on the multiway $V_{proj}$ and $Q_{proj}$ layers in mPLUG-Owl2 and the $V$ and $Q$ matrices in Point-BERT.

\subsection{Comparison with State-of-the-Art Methods}

We selected 15 state-of-the-art PCQA methods for comparison, including 9 FR-PCQA and 5 NR-PCQA methods. 
The FR-PCQA methods are MSE-p2point~\cite{mekuria2016evaluation}, HD-p2point~\cite{mekuria2016evaluation}, MSE-p2plane~\cite{tian2017geometric}, HD-p2plane~\cite{tian2017geometric}, PSNR-yuv~\cite{torlig2018novel}, PointSSIM~\cite{alexiou2020pointssim}, PCQM~\cite{meynet2020pcqm}, MS-GraphSIM~\cite{zhang2021ms}, and MPED~\cite{yang2022mped}.
The NR-PCQA methods are IT-PCQA~\cite{yang2022no}, ResSCNN~\cite{liu2023point}, MM-PCQA~\cite{zhang2022mm}, CoPA+FT~\cite{shan2024contrastive} and LMM-PCQA~\cite{zhang2024lmm}. 
As only one split for LMM-PCQA is available, we reproduce the code and test on our splits.
The other results are reported verbatim from the CoPA+FT paper.

\vspace{-6pt}
\subsubsection{Within-Dataset Performance}

The within dataset performance on LS-PCQA, SJTU-PCQA and WPC is reported in Table~\ref{tab:within_dataset}. From the table, we
observed that our model outperformed all NR-PCQA and FR-PCQA methods on all three datasets. 
Moreover, our model delivered robust performance across all datasets, despite variations in dataset scale, content, and distortion types.
\begin{table*}[t]\small
\centering
\caption{Performance results on the LS-PCQA~\cite{liu2023point}, SJTU-PCQA~\cite{9238424} and WPC~\cite{liu2022perceptual} databases. ``P" and ``I" stand for the the point cloud and image modality, respectively. 
$\uparrow$ indicates that larger is better. 
The best performance results are marked in \textbf{\textcolor{red}{RED}} and the second best results are marked in \textbf{\textcolor{blue}{BLUE}} for both FR-PCQA and NR-PCQA methods. 
``FT'' indicates fine-tuning.}
\begin{tabular}{ccc|cc|cc|cc}
\toprule
    \multirow{2}{*}{Ref}&\multirow{2}{*}{Modal}&\multirow{2}{*}{Methods} & \multicolumn{2}{c|}{LS-PCQA} & \multicolumn{2}{c|}{SJTU-PCQA} & \multicolumn{2}{c}{WPC} \\ 
    \cline{4-9} & & & SROCC $\uparrow$      & PLCC $\uparrow$  & SROCC$\uparrow$      & PLCC$\uparrow$  & SROCC$\uparrow$      & PLCC$\uparrow$ \\ \hline
\multirow{10}{*}{FR} 
 &P&MSE-p2po  & 0.325 & 0.528 & 0.783 & 0.845 & 0.564 & 0.557 \\
 &P&HD-p2po   & 0.291 & 0.488 & 0.681 & 0.748 & 0.106 & 0.166 \\
 &P&MSE-p2pl & 0.311 & 0.498 & 0.703 & 0.779 & 0.445 & 0.491  \\
 &P&HD-p2pl   & 0.291 & 0.478 & 0.617 & 0.661 & 0.344 & 0.380 \\
 &P&PSNR-yuv  & \bf\textcolor{blue}{0.548} & \bf\textcolor{blue}{0.547} & 0.704 & 0.715 & 0.563 & 0.579 \\
  &P&PointSSIM & 0.180 & 0.178 & 0.735 & 0.747  & 0.453 & 0.481 \\
 &P&PCQM      & 0.439 & 0.510 & 0.864 & 0.883 & \bf\textcolor{red}{0.750} & \bf\textcolor{red}{0.754} \\
 &P&MS-GraphSIM & 0.389 & 0.348 & \bf\textcolor{blue}{0.888} & \bf\textcolor{blue}{0.914} & \bf\textcolor{blue}{0.704} & \bf\textcolor{blue}{0.718} \\
 &P&MPED  & \bf\textcolor{red}{0.659} & \bf\textcolor{red}{0.671} & \bf\textcolor{red}{0.898} & \bf\textcolor{red}{0.915} & 0.656 & 0.670 \\ \hline
\multirow{7}{*}{NR} 
 &I&IT-PCQA   & 0.326 & 0.347 & 0.539 & 0.629 & 0.422 & 0.468 \\
 &P&ResSCNN   & 0.594 & 0.624 & 0.834 & 0.863 & 0.735 & 0.752 \\
 &P+I&MM-PCQA   & 0.581 & 0.597 & 0.876 & 0.898  & 0.761 & 0.774 \\
 &P&CoPA+FT     & 0.613 & 0.636  & \bf\textcolor{blue}{0.897} & \bf\textcolor{blue}{0.913} & 0.779 & 0.785 \\
 &P&LMM-PCQA    & \bf\textcolor{blue}{0.684}& \bf\textcolor{blue}{0.691}  & 0.730 & 0.724 & \bf\textcolor{blue}{0.854} & \bf\textcolor{blue}{0.825} \\
 &P+I&\textbf{PIT-QMM} & \bf\textcolor{red}{0.751}& \bf\textcolor{red}{0.766}  & \bf\textcolor{red}{0.906} & \bf\textcolor{red}{0.916} & \bf\textcolor{red}{0.872} & \bf\textcolor{red}{0.844} \\
 
\bottomrule
\end{tabular}
\label{tab:within_dataset}
\end{table*}

\vspace{-6pt}
\subsubsection{Cross-Dataset Performance}

The cross-dataset performance is reported in Table~\ref{tab:cross_dataset}. Since LSPCQA is the largest dataset, followed by WPC, then SJTU, we trained on the full LSPCQA and tested on WPC and SJTU. We also trained on WPC and tested on SJTU. From the Table, it may be observed that PIT-QMM outperforms the other NR-PCQA models, thus demonstrating superior generalizability.

\begin{table}[t]
\renewcommand\tabcolsep{3.3pt}
	\centering
 \vspace{-0.3cm}
  \caption{Cross-dataset evaluation of NR-PCQA methods. Training and testing were both conducted on complete datasets. Results of PLCC are reported.}
  \scriptsize
	\begin{tabular}{cc|ccccc}
		\toprule  
		Train & Test & ResSCNN & MM-PCQA & CoPA+FT & LMM-PCQA & \bf{PIT-QMM} \\ 
		\midrule  
	LS & SJTU & 0.546 & 0.581 & 0.644 & \bf\textcolor{blue}{0.656} & \bf\textcolor{red}{0.682} \\
	LS & WPC & 0.466 & 0.454 & 0.516 & \bf\textcolor{blue}{0.603} & \bf\textcolor{red}{0.648} \\
	WPC & SJTU & 0.572 & 0.612 & \bf\textcolor{blue}{0.643} & 0.597 & \bf\textcolor{red}{0.671} \\
		\bottomrule  
	\end{tabular}
	\label{tab:cross_dataset}
 \vspace{-0.3cm}
\end{table}
\vspace{-6pt}
\subsubsection{Training and Inference Cost}

\begin{table}[t]
\renewcommand\tabcolsep{3.3pt}
	\centering
 \vspace{-0.3cm}
    \caption{Epochs required to converge to best results across all databases. Bold denotes the best performing model.}
  \scriptsize
    \begin{tabular}{c|c|c|c|c}
\toprule
        Method & Batch size & LS-PCQA & SJTU-PCQA & WPC \\\hline
        MM-PCQA & 8 & 50 & \textbf{50} & 50 \\ 
        CoPA + FT & 16 & 20 & 150 & 150 \\
        \bf{PIT-QMM} & 10 & \textbf{5} & 90 & \textbf{30} \\ \bottomrule
    \end{tabular}
    \label{tab:training_cost}
 \vspace{-0.3cm}
\end{table}

As demonstrated in Table~\ref{tab:training_cost}, PIT-QMM converges to the best results when tuning for quality with fewer epochs compared to other SOTA learning-based methods.
The savings were most significant on the large LS-PCQA dataset, where merely 5 epochs were sufficient to obtain SOTA performance.
On the other hand, on the much smaller SJTU-PCQA dataset, we need more epochs, likely as more parameters have to tuned.

PIT-QMM is also efficient for inference, requiring ${\sim}0.9$s per sample of which ${\sim}0.3$s is for preprocessing. This is over 30x faster than LMM-PCQA for a cloud of 1 million points, which involves expensive handcrafted feature extraction.


\subsection{Ablation Study}
\label{sec:ablation}

\begin{table}[t]\small
\renewcommand\tabcolsep{3.3pt}
\centering
\vspace{-0.3cm}
\caption{Ablation study on the LSPCQA-small~\cite{liu2023point} and WPC~\cite{liu2022perceptual} databases. 
$\uparrow$ indicates that larger is better.}
\begin{tabular}{c|cc|cc}
\toprule
    \multirow{2}{*}{Methods} & \multicolumn{2}{c|}{LSPCQA-small} & \multicolumn{2}{c}{WPC} \\ 
    \cline{2-5}
    & SROCC$\uparrow$      & PLCC$\uparrow$  & SROCC$\uparrow$      & PLCC$\uparrow$ \\ \hline
 \textcircled{1} & 0.684 & 0.664 & 0.837 & 0.804 \\
 \textcircled{2} & 0.722 & 0.681 & 0.866 & 0.835 \\
 \textcircled{3} & 0.734 & 0.699 & 0.872 & 0.844  \\
 \textcircled{4} & 0.730 & 0.694 & 0.865 & 0.839 \\
 \textcircled{5} & 0.343 & 0.322 & 0.447 & 0.405 \\
 \textcircled{6} & 0.733 & 0.704 & 0.870 & 0.832 \\
 \textcircled{7} & 0.737 & 0.706 & 0.869 & 0.838 \\
\bottomrule
\end{tabular}
\label{tab:ablation}
\end{table}	

We conducted an ablation study to evaluate the contributions of different components in our proposed dataset construction strategy. 
Table~\ref{tab:ablation} summarizes the results of this study. We used WPC and LSPCQA-small databases in these ablations.

First, using only 2D image projections to predict quality (row \textcircled{1}) yielded strong performance on both datasets, validating the use of pretrained vision models. However, performance improved when point cloud data was incorporated.

Next, we examined three point cloud sampling schemes: local patches (row \textcircled{2}), adding furthest point samples (row \textcircled{3}), and multi-scale sampling with half-scale patches (row \textcircled{4}). Sampling local patches alone showed limited improvement due to the pretrained encoder's domain gap, which focuses on semantic understanding of object-like point clouds. Adding furthest point samples improved results by introducing content-oriented features. However, incorporating multi-scale information had minimal effect. Likely, the patches need to be matched before processing, so that the encoders would become receptive to the fine details.

Using only point cloud features (row \textcircled{5}) significantly decreased performance, highlighting the domain gap in pretrained encoders. Lastly, varying text prompts with additional task, psychometric (row \textcircled{6}), and rendering contexts (row \textcircled{7}) slightly improved performance.
\vspace{-10pt}
\subsection{Distortion Identification and Localization}

We report the result of our localization experiments in Table~\ref{tab:localize}. Since there are no existing baselines for this task, we compared against a ViT and a Q-Align model trained to predict the category and the octant of distortion from cubic projections. Synthetic data was generated from LSPCQA-small. First, we observe that the ViT baseline performed poorly on this task, likely due to a significant domain shift. Next, we observe that Q-Align also demonstrated strong localization abilities, which is expected for an LMM-based method. Finally, PIT-QMM outperformed both baselines with the help of the view-based positional embeddings and point cloud features.

\begin{table}[t]
\renewcommand\tabcolsep{3.3pt}
	\centering
 \vspace{-0.3cm}
    \caption{Accuracy on distortion identification and localization tasks. Bold denotes the best performing model.}
  \scriptsize
    \begin{tabular}{c|c|c}
\toprule
        Method & Identification Acc. & Localization Acc. \\\hline
        ViT & 53.8\% & 28.1\%  \\
        Q-Align & 79.1\% & 72.7\%  \\ 
        \bf{PIT-QMM} & \bf84.3\% & \bf75.2\%  \\ \bottomrule
    \end{tabular}
    \label{tab:localize}
 \vspace{-0.3cm}
\end{table}

\section{Conclusion}

In this paper, we presented a novel end-to-end LMM-based NR-PCQA algorithm. By leveraging complementary information from different modalities and large pretrained encoders, our proposed PIT-QMM model predicts quality scores across a wide variety of distortion and content types. Extensive experiments show that PIT-QMM achieves competitive performance across varied benchmarks with fewer training iterations than other SOTA models. Preliminary experiments show that PIT-QMM can also pinpoint the nature and location of distortions with high accuracy, which indicates an exciting new path towards interactive and explainable quality agents.



\bibliographystyle{IEEEbib}
\bibliography{strings,refs}

\begin{thebibliography}{10}

\bibitem{zhang2022pointclip}
R.~Zhang, Z.~Guo, W.~Zhang, K.~Li, X.~Miao, B.~Cui, et~al.,
\newblock ``Pointclip: Point cloud understanding by clip,''
\newblock in {\em IEEE/CVF Conference on Computer Vision and Pattern Recognition}, 2022, pp. 8552--8562.

\bibitem{wang2004image}
Z.~Wang, A.~C. Bovik, H.~R. Sheikh, and E.~P. Simoncelli,
\newblock ``Image quality assessment: from error visibility to structural similarity,''
\newblock {\em IEEE Transactions on Image Processing}, vol. 13, no. 4, pp. 600–612, Apr. 2004.

\bibitem{liu2023point}
Y.~Liu, Q.~Yang, Y.~Xu, and L.~Yang,
\newblock ``Point cloud quality assessment: Dataset construction and learning-based no-reference metric,''
\newblock {\em ACM Transactions on Multimedia Computing, Communications and Applications}, vol. 19, no. 2s, pp. 1--26, 2023.

\bibitem{zhang2022mm}
Z.~Zhang, W.~Sun, X.~Min, Q.~Zhou, J.~He, Q.~Wang, and G.~Zhai,
\newblock ``Mm-pcqa: Multi-modal learning for no-reference point cloud quality assessment,''
\newblock {\em arXiv preprint arXiv:2209.00244}, 2022.

\bibitem{shan2024contrastive}
Z.~Shan, Y.~Zhang, Q.~Yang, H.~Yang, Y.~Xu, J.~Hwang, X.~Xu, and S.~Liu,
\newblock ``Contrastive pre-training with multi-view fusion for no-reference point cloud quality assessment,''
\newblock in {\em IEEE/CVF Conference on Computer Vision and Pattern Recognition}, 2024, pp. 25942--25951.

\bibitem{wu2023q}
H.~Wu, Z.~Zhang, W.~Zhang, et~al.,
\newblock ``Q-align: Teaching lmms for visual scoring via discrete text-defined levels,''
\newblock {\em arXiv preprint arXiv:2312.17090}, 2023.

\bibitem{zhang2024lmm}
Z.~Zhang, H.~Wu, Y.~Zhou, C.~Li, W.~Sun, et~al.,
\newblock ``Lmm-pcqa: Assisting point cloud quality assessment with lmm,''
\newblock {\em arXiv preprint arXiv:2404.18203}, 2024.

\bibitem{Point-BERT}
X.~Yu, L.~Tang, Y.~Rao, T.~Huang, J.~Zhou, and J.~Lu,
\newblock ``Point-bert: Pre-training 3d point cloud transformers with masked point modeling,''
\newblock in {\em CVPR}, 2022.

\bibitem{Cap3D}
T.~Luo, C.~Rockwell, H.~Lee, and J.~Johnson,
\newblock ``Scalable 3d captioning with pretrained models,''
\newblock {\em arXiv:2306.07279}, 2023.

\bibitem{hu2021lora}
E.~J. Hu, Y.~Shen, P.~Wallis, Z.~Allen-Zhu, et~al.,
\newblock ``Lora: Low-rank adaptation of large language models,''
\newblock {\em arXiv preprint arXiv:2106.09685}, 2021.

\bibitem{9238424}
Q.~Yang, H.~Chen, Z.~Ma, Y.~Xu, R.~Tang, and J.~Sun,
\newblock ``Predicting the perceptual quality of point cloud: A 3d-to-2d projection-based exploration,''
\newblock {\em IEEE Transactions on Multimedia}, vol. 23, pp. 3877--3891, 2021.

\bibitem{liu2022perceptual}
Q.~Liu, H.~Su, Z.~Duanmu, W.~Liu, and Z.~Wang,
\newblock ``Perceptual quality assessment of colored 3d point clouds,''
\newblock {\em IEEE Transactions on Visualization and Computer Graphics}, vol. 29, no. 8, pp. 3642--3655, 2022.

\bibitem{xue2024ulip}
L.~Xue, N.~Yu, S.~Zhang, A.~Panagopoulou, J.~Li, R.~Mart{\'\i}n-Mart{\'\i}n, J.~Wu, C.~Xiong, et~al.,
\newblock ``Ulip-2: Towards scalable multimodal pre-training for 3d understanding,''
\newblock in {\em IEEE/CVF Conference on Computer Vision and Pattern Recognition}, 2024, pp. 27091--27101.

\bibitem{xu2023pointllm}
R.~Xu, X.~Wang, T.~Wang, Y.~Chen, J.~Pang, and D.~Lin,
\newblock ``Pointllm: Empowering large language models to understand point clouds,''
\newblock {\em arXiv preprint arXiv:2308.16911}, 2023.

\bibitem{mekuria2016evaluation}
R.~Mekuria, Z.~Li, C.~Tulvan, and P.~Chou,
\newblock ``Evaluation criteria for point cloud compression,''
\newblock {\em ISO/IEC MPEG}, 2016.

\bibitem{tian2017geometric}
D.~Tian, H.~Ochimizu, C.~Feng, R.~Cohen, and A.~Vetro,
\newblock ``Geometric distortion metrics for point cloud compression,''
\newblock in {\em IEEE International Conference on Image Processing}, 2017, pp. 3460--3464.

\bibitem{torlig2018novel}
E.~M. Torlig, E.~Alexiou, T.~A. Fonseca, R.~L. de~Queiroz, and T.~Ebrahimi,
\newblock ``A novel methodology for quality assessment of voxelized point clouds,''
\newblock in {\em Applications of Digital Image Processing XLI}, 2018, vol. 10752, pp. 174--190.

\bibitem{alexiou2020pointssim}
E.~Alexiou and T.~Ebrahimi,
\newblock ``Towards a point cloud structural similarity metric,''
\newblock in {\em IEEE International Conference on Multimedia and Expo Workshops}, 2020, pp. 1--6.

\bibitem{meynet2020pcqm}
G.~Meynet, Y.~Nehm{\'e}, J.~Digne, and G.~Lavou{\'e},
\newblock ``Pcqm: A full-reference quality metric for colored 3d point clouds,''
\newblock in {\em International Conference on Quality of Multimedia Experience}, 2020, pp. 1--6.

\bibitem{zhang2021ms}
Y.~Zhang, Q.~Yang, and Y.~Xu,
\newblock ``Ms-graphsim: Inferring point cloud quality via multiscale graph similarity,''
\newblock in {\em ACM International Conference on Multimedia}, 2021, pp. 1230--1238.

\bibitem{yang2022mped}
Q.~Yang, Y.~Zhang, S.~Chen, Y.~Xu, J.~Sun, and Z.~Ma,
\newblock ``Mped: Quantifying point cloud distortion based on multiscale potential energy discrepancy,''
\newblock {\em IEEE Transactions on Pattern Analysis and Machine Intelligence}, vol. 45, no. 5, pp. 6037--6054, 2022.

\bibitem{yang2022no}
Q.~Yang, Y.~Liu, S.~Chen, Y.~Xu, and J.~Sun,
\newblock ``No-reference point cloud quality assessment via domain adaptation,''
\newblock in {\em IEEE/CVF Conference on Computer Vision and Pattern Recognition}, 2022, pp. 21179--21188.

\end{thebibliography}

\newpage
\appendix
\section{Appendix}

\subsection{Psychometric Setup Description in Prompt}

We hypothesized that including details of the psychometric experiment in the prompt might guide the model towards better predictions. As an example, we included the following description from LS-PCQA when training and testing on it -- 
\begin{quote}
    In the subjective experiment, the participants sit in a controlled environment. Specifically, the zoom rate is set as 1:1. The presentation device used in subjective experiments is Dell SE2216H with a 21.5-inch monitor with a resolution of 1920×1080 pixels. The sitting posture of the participants is adjusted to ensure that their eyes are at the same height as the center of the screen. The viewing distance is about three times the height of the rendered point cloud ($\approx 0.75$ meters). The subjective experiment is conducted indoors, under a normal lighting condition.
\end{quote}
We included a similar description for other datasets as available. Row \textcircled{6} in Table~\ref{tab:ablation} shows the effect of adding this psychometric context. 
We see a slight improvement in our evaluation metrics, but the performance is comparable with the task only prompt (row \textcircled{5}).
We believe this is likely as the LLM is already able to draw this information as relevant world knowledge from the task section of the prompt and does not particularly need further explicit details.

\subsection{Effect of Rendering Parameters on Perceptual Quality}

\begin{figure}[h]
    \centering
    \includegraphics[width=0.8\linewidth]{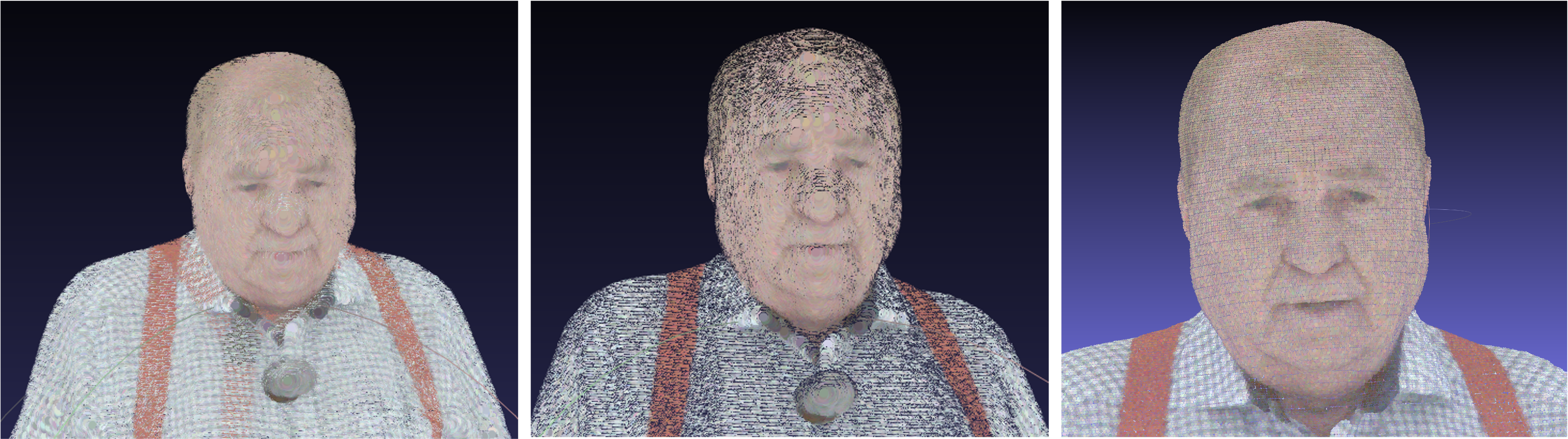}
    \caption{The same underlying point cloud can have highly different quality characteristics depending on rendering parameters and the radius of interaction, especially in the NR setting. Point cloud taken from LS-PCQA and rendered in MeshLab. Best viewed zoomed in.}
    \label{fig:rendering}
\end{figure}

We observed that quality assessment for point clouds is highly dependent on the settings used to render the point cloud and how the user was allowed to interact with it. 
For example, Figure~\ref{fig:rendering} shows the same point cloud rendered with different point sizes and viewing distances, all of which have significantly different quality characteristics.
This is a complexity typically not observed in 2D quality datasets.
Accordingly, we added rendering parameters in our prompt as described in the corresponding datasets when available or a best effort reproduction when not.
Method \textcircled{7} in Table~\ref{tab:ablation} shows the effect of including these parameters.
As an example, we added the following description for LS-PCQA -- 
\begin{quote}
    The point cloud is rendered with a point size of 2 mm with cameras at 2.5m from the object and perspective projection with square primitives.
\end{quote}
The improvement is modest over the base case. We believe this is likely because this information can be inferred from a combination of the image projections and the text description of the task, so specifying it explicitly has relatively little impact.

\subsection{Further Implementation Details}

The point cloud projections were rendered with PyTorch3D at a resolution of $512\times 512$. 
All point cloud samples are $n=8192$ dimensional with 3 spatial coordinates and 3 RGB color coordinates, which makes $d=6$. 
The furthest point sampling was done with the Python package fpsample with the bucket-based FPS algorithm. 
To sample local patches, we constructed a search tree using the Python package FAISS, sampled a single point randomly and then looked up the closest points near it to construct the final sample. 
For the two scale patching, uniform downsampling is conducted with Open3D at a factor of 2.
The point encoder outputs $m = 513$ point features, each with $c = 384$ dimensions.
The point feature projector contains three linear layers with the GeLU activation, which maps point features to tokens with $c' = 5120$ dimensions.
Since we added two additional special tokens, the vocabulary size of PIT-QMM is $V = 32003$.
The weights of the image encoder and LLM are initialized from Q-Align.

\subsection{On Training Efficiency}

We report the number of epochs for each model in Table~\ref{tab:training_cost} verbatim from the respective technical reports or the code provided. 
A subtlety in this comparison is that the batch size for all of these models are different, so overall training iterations would vary.
However, the batch sizes are within the same range (8-20), so the trends should remain similar even after batch size is normalized.
Note that the batch size we used for PIT-QMM is relatively low, so normalizing for a larger batch size as used elsewhere would likely favour our model.

\end{document}